\documentclass[journal,a4paper]{IEEEtran}

\usepackage{graphicx}

\usepackage[overlay,absolute]{textpos}

\begin{document}

\begin{textblock*}{10in}(28mm, 8mm)
{\textbf{Ref:} \emph{IEEE Transactions on Evolutionary Computation}, Vol. 18, No. 5, pp. 779-789, September 2014.}
\end{textblock*}

\begin{textblock*}{10in}(33mm, 13mm)
{\fbox{Winner of Gold Award in 11th Annual ``Humies'' Awards for Human-Competitive Results }}
\end{textblock*}

\author{Eli (Omid) David$^1$,
        H.~Jaap van den Herik$^2$,
        Moshe Koppel$^3$,
		Nathan S.~Netanyahu$^4$
\thanks{$^1$ Department of Computer Science, Bar-Ilan University, Ramat-Gan 52900, Israel (mail@elidavid.com, www.elidavid.com).}% <-this % stops a space
\thanks{$^2$ Tilburg Center for Cognition and Communication (TiCC), Tilburg University, Tilburg, The Netherlands (h.j.vdnherik@uvt.nl).}% <-this % stops a space
\thanks{$^3$ Department of Computer Science, Bar-Ilan University, Ramat-Gan 52900, Israel (koppel@cs.biu.ac.il).}
\thanks{$^4$ Department of Computer Science, Bar-Ilan University, Ramat-Gan 52900, Israel (nathan@cs.biu.ac.il), and Center for Automation Research, University of Maryland, College Park, MD 20742 (nathan@cfar.umd.edu).}
}

\title{Genetic Algorithms for Evolving\\ Computer Chess Programs}

\maketitle

\begin{abstract}

This paper demonstrates the use of genetic algorithms for evolving (1) a grandmaster-level evaluation function and (2) a search mechanism for a chess program, the parameter values of which are initialized randomly. The evaluation function of the program is evolved by learning from databases of (human) grandmaster games. At first the organisms are evolved to mimic the behavior of human grandmasters, and then these organisms are further improved upon by means of coevolution. The search mechanism is evolved by learning from tactical test suites. Our results show that the evolved program outperforms a two-time World Computer Chess Champion, and is on a par with other leading computer chess programs. 

\end{abstract}


\section{Introduction}

Despite the many advances in Machine Learning and Artificial Intelligence, there are still areas where learning machines have not yielded a performance comparable to the top performance exhibited by humans. Computer Chess is one of the most difficult areas with this aim.

It is well known that computer games have served as an important testbed for spawning various innovative AI techniques in domains and applications such as search, automated theorem proving, planning, and learning. In addition, the annual World Computer Chess Championship (WCCC) is arguably the longest ongoing performance evaluation of programs in computer science, which has inspired other well known competitions in robotics, planning, and natural language understanding.

Computer chess, while being one of the most researched fields within AI, has not lent itself to the successful application of conventional learning methods, due to its enormous complexity. Hence, top chess programs still resort to manual tuning of the parameters of their evaluation function and their search functions, typically through years of trial and error. The evaluation function assigns a score to a given chess position, and the selective search function decides which moves to search more deeply and which moves to prune at an earlier stage in the search tree.

Our previous work described how the parameters of an evaluation function can be evolved by learning from other chess programs \cite{david11} and human chess players \cite{david09}. It also provided a method for tuning the search parameters of a chess program \cite{david10}. That is, we developed a method for evolving the parameters of a chess program's evaluation function given a tuned search mechanism, and a method for evolving the parameters of the search mechanism given a highly tuned evaluation function. 

In this paper we extend our previous work by providing a methodology for evolving the parameters of both the evaluation function and the selective search mechanism of a chess program using Genetic Algorithms (GA). That is, we assume no access to either a tuned function or a tuned search mechanism, but rather initialize the parameters of both components randomly (i.e., the initial program includes only the rules of the game, a set of randomly initialized evaluation function parameters and search function parameters). Additionally, we provide extensive experimental results demonstrating that the combined approach is capable of evolving highly tuned chess programs, on a par with top tournament-playing chess programs.

We observe that although the experimental results of the resulting approach presented in this paper, in terms of the evolved parameter values, the number of positions solved using a test suite, and the results of matches against other top programs differ from those of our previous work, the overall performances are comparable. This suggests that the algorithm is advantageous in the sense that it initially relies only on databases of (human) grandmaster games.

At the first stage we evolve the parameters of the evaluation function, relying on a straightforward mechanism for conducting 1-ply searches only. That is, since we do not have yet a sophisticated selective search mechanism, we use a plain mechanism for making legal 1-ply moves. After evolving the parameters of the evaluation function and incorporating its evolved values, we evolve at the second stage the parameters of our selective search mechanism.

In Section \ref{sec:learning} we review past attempts at applying evolutionary techniques in computer chess. We also compare alternative learning methods to evolutionary methods, and argue why the latter are more appropriate for evolving an evaluation function. Section \ref{sec:evolution} presents our new approach, including a detailed description of the framework of the GA as applied to evolving the parameters of a chess program's evaluation function. Section \ref{sec:search-evolution} provides a GA-based method for tuning the parameter values of the selective search mechanism of the program. Section \ref{sec:experiments} provides our extended experimental results, and Section \ref{sec:conclusions} contains concluding remarks.

\section{\label{sec:learning}Learning in Computer Chess}

While the first chess programs could not pose a challenge to even a novice player \cite{shannon50,turing53}, the current advanced chess programs have been outperforming the strongest human players, as the recent man vs.~machine matches clearly indicate \cite{wullenweber06}. This improvement is largely a result of deep searches that are possible nowadays, thanks to both hardware speed and improved search techniques. While the search depth of early chess programs was limited to only a few plies, nowadays tournament-playing programs easily search more than a dozen plies in the middlegame, and tens of plies in the late endgame.

Despite their groundbreaking achievements, a glaring deficiency of today's top chess programs is their severe lack of a learning capability (except in most negligible ways, e.g., ``learning'' not to play an opening that resulted in a loss, etc.). In other words, despite their seemingly intelligent behavior, those top chess programs are mere brute-force (albeit efficient) searchers that lack true underlying intelligence.

\subsection{Conventional vs. Evolutionary Learning in Computer Chess}

During more than fifty years of research in the area of computer games, many learning methods have been employed in less complex games. Chellapilla and Fogel \cite{chellapilla99a,chellapilla99b,chellapilla01,fogel02} created an expert-level checkers program by using coevolution to evolve neural network board evaluators. Temporal difference learning has been successfully applied in backgammon \cite{tesauro92} and checkers \cite{schaeffer01}. Although temporal difference learning has also been applied to chess \cite{baxter00}, the results showed that after three days of learning, the playing strength of the program was merely 2150 Elo (see Appendix B for a description of the Elo rating system), which is quite a low rating for a chess program. These experimental results confirmed Wiering's \cite{wiering95} formal arguments, presented five years earlier, for the failure of these methods in rather complex games such as chess. Runarsson and Lucas \cite{runarsson12} compared least squares temporal difference learning (LSTD($\lambda$)) with preference learning in the game of Othello, using samples of games from human competitions held by the French Othello Federation. Their results showed that preference learning produces policies that better capture the behavior of expert players, and also lead to higher levels of play when compared to LSTD($\lambda$). It would be interesting to apply this approach to chess, which is more complex than Othello. For example, while the average branching factor (i.e., the average number of legal moves available in each position) in chess is 38, it is 7 in Othello, and smaller than 3 in checkers (when considering both capture and non-capture positions).

The issue of learning in computer chess is basically an optimization problem. Each program plays by conducting a search, where the root of the search tree is the current position, and the leaf nodes (at some predefined depth of the tree) are evaluated by some static evaluation function. In other words, sophisticated as the search algorithms may be, most of the knowledge of the program lies in its evaluation function. Even though automatic tuning methods, that are based mostly on reinforcement learning, have been successfully applied to less complex games such as checkers, they have had almost no impact on state-of-the-art chess engines. Currently, all top tournament-playing chess programs use hand-tuned evaluation functions, since conventional learning methods cannot cope with the enormous complexity of the problem. This is underscored by the following four points.

(1) The space to be searched is huge. It is estimated that there are about $10^{46}$ different positions possible that can arise in chess \cite{chinchalkar96}. As a result, any method based on exhaustive search of the problem space is so far infeasible.

(2) The search space is not smooth and unimodal. The evaluation function's parameters of any top chess program are highly co-dependent. An example from the first author's computer chess career \cite{david05,david09a} may illustrate this. In many cases increasing the values of three parameters results in a worse performance, but if a fourth parameter were also increased, then an improved overall performance would be obtained. Since the search space is not unimodal, i.e., it does not consist of a single smooth ``hill'', any gradient-ascent algorithm such as hill climbing will perform poorly. In contrast, genetic algorithms are known to perform well in large search spaces which are not unimodal.

(3) The problem of tuning and learning is not well understood. As will be discussed in detail in the next section, even though all top programs are hand-tuned by their programmers, finding the best value for each parameter is based mostly on educated guessing and intuition. (The fact that all top programs continue to operate in this manner attests to the lack of practical alternatives.) Had the problem been well understood, a domain-specific heuristic would have outperformed a general-purpose method such as GA.

(4) We do not require a global optimum to be found. Our goal in tuning an evaluation function is to adjust its parameters so that the overall performance of the program is enhanced. In fact, a unique global optimum most probably does not exist for this tuning problem.

In view of the above four points it seems appropriate to employ GA for automatic tuning of the parameters of an evaluation function. Indeed, at first glance this appears like an optimization task, well suited for GA. The many parameters of the evaluation function (bonuses and penalties for each property of the position) can be encoded as a bit-string. We can randomly initialize many such ``chromosomes'', each representing one evaluation function. Thereafter, one needs to evolve the population until highly tuned ``fit'' evaluation functions emerge.

However, there is one major obstacle that hinders the above application of GA, namely the fitness function. Given a set of parameters of an evaluation (encoded as a chromosome), how should the fitness value be calculated? A straightforward approach would let the chromosomes in each generation play against each other a series of games, and subsequently, record the score of each individual as its fitness value. (Each ``individual'' is a chess program with an appropriate evaluation function.)

The main drawback of this approach is the large amount of time needed to evolve each generation. As a result, limitations should be imposed on the length of the games played after each generation, and also on the size of the population involved. With a population size of 100, a 10 seconds limit per game, and assuming that each individual plays each other individual once in every generation, it would take 825 minutes for each generation to evolve. Specifically, reaching the 100th generation would take up to 57 days. These figures suggest that it would be rather difficult to evolve the parameter values of a chess program relying on coevolution alone. (When some a priori knowledge regarding material and positional values is used, as in Fogel \emph{et al.}~\cite{fogel04,fogel05,fogel06} for example, good results are attainable in reasonable time.)

In Section \ref{sec:evolution} we present our approach for using GA in evolving state-of-the-art chess evaluation functions. Before that, we briefly review previous work in applying evolutionary methods in computer chess.

\subsection{Previous Evolutionary Methods Applied to Chess}

Despite the abovementioned problems, there have been some successful applications of evolutionary techniques in computer chess, subject to some restrictions. Genetic programming was successfully employed by Hauptman and Sipper \cite{hauptman05,hauptman07} for evolving programs that can solve Mate-in-N problems and play chess endgames.

Kendall and Whitwell \cite{kendall01} used evolutionary algorithms for tuning the parameters of an evaluation function. Their approach had limited success, due to the very large number of games required (as previously discussed), and the small number of parameters used in their evaluation function. Their evolved program managed to compete with strong programs only if their search depth (lookahead) was severely limited.

Similarly, Aksenov \cite{aksenov04} employed genetic algorithms for evolving the parameters of an evaluation function, using games between the organisms for determining their fitness. Again, since this method required a very large amount of games, it evolved only a few parameters of the evaluation function with limited success. Tunstall-Pedoe \cite{tunstall91} also suggested a similar approach, without providing an implementation.

Gross \emph{et al.}~\cite{gross02} combined genetic programming and evolution strategies to improve the efficiency of a given search algorithm using a distributed computing environment on the Internet.

Fogel \emph{et al.}~\cite{fogel04,fogel05,fogel06} used coevolution to successfully improve the parameters of an existing chess program. Their algorithm learns to evaluate chessboard configurations by using the positions of pieces, material and positional values, and neural networks to assess specific sections of the chessboard. The method succeeded in modifying the parameter values of an existing chess program to gain a respectable performance level (the program scored 3 wins, 11 draws, and 10 losses against \textsc{Fritz 8} and defeated a human master). As far as we know, no successful attempt has been described at using coevolution to evolve the parameters of a chess program from fully randomized initial values, without relying on any a priori knowledge. Fogel \emph{et al.} showed that coevolution could be employed successfully in chess when the initial material and positional parameters are already initialized within sensible ranges. Furthermore, Chellapilla and Fogel \cite{chellapilla99a,chellapilla99b,chellapilla01,fogel02} successfully employed coevolution to evolve the parameters of a checkers program. However, the applicability of their approach to chess is not clear, as the game of chess is by orders of magnitude more complex than checkers. (The fact that checkers is by now a solved game \cite{schaeffer07} further attests to its relative simplicity in comparison to chess, which seems far from being solved.)

Samothrakis \emph{et al.}~\cite{samothrakis12} used covariance matrix adaptation evolution strategy (CMA-ES) for coevolution of weighted piece counters (WPCs) in Othello. These WPCs were used to operate as value functions in a 1-ply minimax player. The aim was not to find Othello players that are strong in absolute terms, but rather use Othello as an interesting domain of study in which to measure performance and intransitivities in coevolution. Using CMA-ES within a coevolutionary setting, the authors succeeded in evolving the strongest WPCs for Othello, published as of yet. While this coevolutionary approach yields interesting results for Othello, it would be challenging to apply it to chess, which is more complex than Othello (as noted in the previous section).

There are two main components of any chess program, an evaluation function and a search mechanism. In our previous work, we provided a ``mentor-assisted'' approach \cite{david11} for reverse engineering the evaluation function of a target chess program (the ``mentor''), thereby evolving a new comparable evaluation function. This approach relied on the fact that the mentor provides an evaluation score for any given position, and so, using a large set of positions, the evaluation score of the target program can be obtained for each position. Subsequently, the parameters of the evaluation function can be evolved to mimic the scores provided by the mentor. This mentor-assisted approach produces an evaluation function that mimics a given chess program, but is dependent on access to the evaluation score of the target program.

In order to extend the concept of mentor-assisted evolution to learning from humans as well, we combined evolution and coevolution for evolving the parameter values of the evaluation function to simulate the moves of a human grandmaster, without relying on the availability of evaluation scores of some computer chess programs \cite{david09}. In both these versions we had assumed that the program already contained a highly tuned search mechanism.

While most past attempts at automatic learning in chess focused on an evaluation function, few efforts have been made to evolve automatically the parameters controlling the search mechanism of the program. Moriarty and Miikkulainen \cite{moriarty94} used neural networks for tuning the search of an Othello program, but as mentioned in their paper, their method is not easily applicable to more complex games such as chess. Cazenave \cite{cazenave01} presented a method for learning search-control rules in Go, but the method cannot be applied to chess (the search mechanism in Go is of a different, less critical nature than chess).

Bj\"ornsson and Marsland \cite{bjornsson02} presented a method for automatically tuning search extensions in chess. Given a set of test positions (for which the correct move is predetermined) and a set of parameters to optimize (in their case, four extension parameters), they tune the values of the parameters using gradient descent. Their program processes all the positions, and for each position it records the number of nodes until a solution is found. The optimization goal is to minimize the total node count for all the positions. In each iteration, their method modifies each of the extension parameters by a small value, and processes all the positions, recording the total node count. Thus, given $N$ parameters to be optimized ($N=4$ in their case), in each iteration their method processes all the positions $N$ times. The parameter values are updated after each iteration, such that the total node count decreases. Bj\"ornsson and Marsland applied their method for tuning the values of four search extension parameters: check, passed pawn, recapture, and one-reply extensions. Their results show that their method optimizes the values for these parameters such that the total node count for solving the test set is decreased. 

Despite the success of this gradient-descent method for tuning the values of four search extension parameters, it cannot be used efficiently to optimize a large set of parameters consisting of all the critical selective search parameters, of which search extensions comprise only a few parameters. This is due to the fact that the gradient-descent approach requires processing the whole test set for each of the parameters in each iteration. While this might be practical for four parameters, it becomes more difficult when the number of parameters involved is considerably larger. Additionally, unlike the optimization of search extensions, where parameter values are mostly independent, other search methods would exhibit a high interdependency between the parameter values, and consequently it would be more difficult to apply gradient-descent optimization.

In our previous work we demonstrated \cite{david10} that given a chess program with a highly tuned evaluation function, we can use GA to evolve the search parameters by evolving the parameters such that the overall performance in test suites is improved. As mentioned previously, we have also demonstrated \cite{david09,david11} how to evolve an evaluation function using GA, given a highly tuned search mechanism.

In this paper we combine the two approaches in order to evolve both the evaluation function and the search mechanism, assuming that the parameters of none of these two components are tuned a priori (but rather initialized randomly).

\section{\label{sec:evolution}Evolution and Coevolution of Evaluation Functions}

The parameters of an evaluation function can easily be represented as a chromosome. However, applying a fitness function is more complicated. As previously noted, establishing a fitness evaluation by means of playing numerous games between the organisms in each generation (i.e., single-population coevolution) is quite difficult (when the organisms are initialized randomly). 

In our previous work on mentor-assisted evolution \cite{david11} we described how the fitness value can be issued by running a grandmaster-level chess program on a set of positions, and recording for each position the difference between the evaluation score computed by the organism and the score computed by the target program. We define (1) this difference as the \emph{evaluation error} and (2) a magnitude inversely proportional to it as the \emph{fitness function}. However, this approach is not practical when trying merely to learn from a player's moves (whether a human player or a computer program, assuming no access to the program itself).

This paper extends significantly our previous framework. Specifically, we present a learning approach which relies only on widely available databases of grandmaster-level games. This task is significantly more difficult than using existing chess programs as mentors, since for any position taken from a (human) game, the only available information is the move actually played (and not the associated score). Our approach is based on the steps shown in Figure \ref{fig:mentor}.

\begin{figure}[htbp]
\begin{center}
\line(1,0){240}

\begin{enumerate}
\item Using a database of grandmaster-level games, select a list of positions and the move played in that position.
\item For each position, let the organism perform a 1-ply search and store the move selected by the organism.
\item Compare the move suggested by the organism with the actual move made by the player. The fitness of the organism will be the total number of moves where the organism's move is the same as the player's move.
\end{enumerate}

\line(1,0){240}
\caption{Fitness function for evolution of evaluation functions using database of grandmaster-level games.}
\label{fig:mentor}
\end{center}
\end{figure}

Although performing a search for each position seems to be a costly process, in fact it consumes little time. Conducting a 1-ply search amounts to less than a millisecond for a typical chess program on an average machine, and so one thousand positions can be processed in one second. This allows us to use a large set of positions for the training set.

After running the abovementioned process a number of times, we obtain several evolved organisms. Due to the random initialization of the chromosomes, each time the process is applied, a different ``best evolved organism'' is obtained. Comparing the best evolved organisms from different runs, we observe that even though they are of similar playing strength, their evolved parameter values differ, and so does their playing style.

In order to further improve the obtained organisms (each organism is the best evolved organism from a complete run of the abovementioned process), we next use a single-population coevolution phase. During this phase the evolved organisms play against each other, and the fitness function applied is based on this relative performance. After running this coevolution for a predetermined number of generations, the best evolved organism is selected as the best overall organism. 

As the results in Section~\ref{sec:experiments} indicate, this ``best of best'' organism improves upon the organisms evolved from the evolutionary phase only. As noted before, it is difficult to evolve the parameter values of a chess program from randomly initialized values relying on coevolution alone. The difference in our case is that the population size is small (we used 10), and the initial organisms are already well tuned, rather than randomly initialized.

In the following subsections, we describe in detail the chess program, the implementation of the evolutionary method, and the GA parameters used.

\subsection{The Chess Program and the Evaluation Function}

The evaluation function of our program (which we are interested in tuning automatically) consists of 35 parameters (see Figure~\ref{fig:best-evolved}). Even though this is a small number of parameters in comparison to other top programs, the set of parameters used does cover all important aspects of a position, e.g., material, piece mobility and centricity, pawn structure, and king safety. 

The parameters of the evaluation function are represented as a binary bit-string (chromosome size: 224 bits), initialized randomly. (Note that we use binary encoding since it is the most basic type of encoding for GA \cite{holland75}, although it is not necessarily superior to an alternative encoding method for this problem.) The value of a pawn is set to a fixed value of 100, which serves as a reference for all other parameter values. Except for the four parameters representing the material values of the pieces, all the other parameters are assigned a fixed length of 6 bits per parameter. Obviously, there are many parameters for which 3 or 4 bits suffice. However, allocating a fixed length of 6 bits to all parameters ensures that a priori knowledge does not bias the algorithm in any way. Note that at this point, the program's evaluation function is merely a random initialization, so that apart from the rules of the game, the program has essentially no game skills at all.

\vspace{4pt}

\subsection{Evolution using Grandmaster Games}

For our experiments, we used a database of 10,000 games of grandmasters rated above 2600 Elo, and randomly picked one position from each game. We picked winning positions only, i.e., positions where the side to move had ultimately won the game (e.g., if it was white's turn to move, the game would be won eventually by white). Of these 10,000 positions, we selected 5,000 positions for training and 5,000 for testing.

In each generation, for each organism the algorithm translates its chromosome bit-string to a corresponding evaluation function. For each of the $N$ test positions (in our case, $N=5,000$), it then performs a 1-ply search using the decoded evaluation function, and the best move returned from the search is compared to that of the grandmaster in the actual game. The move is deemed ``correct'' if it is the same as the move played by the grandmaster, and ``incorrect'' otherwise. The fitness of the organism is calculated as the square of the total number of correct moves.

Note that unlike the mentor-assisted approach for mimicking existing chess programs, which provide numeric values for each position, here we only have 1-bit of information for each processed position (correct/incorrect). This underscores why learning merely from a player's moves is much more challenging than using computer programs as mentors.

Other than the special fitness function described above, we used a standard GA implementation with Gray coded chromosomes, fitness-proportional selection, uniform crossover, and elitism (the best organism is copied to the next generation). The following parameters are used by the algorithm:
\\
\\
population size = 100,\\
crossover rate = 0.75,\\
mutation rate = 0.005,\\
number of generations = 200.

\subsection{Coevolution of the Best Evolved Organisms}

We ran the evolution process ten times, thus obtaining ten organisms, each being the best organism from one run. The parameter values of these ten organisms differ due to the random initialization in each run. Consequently, although these ten programs are of similar playing strength, their playing style is different. Note that using the top ten evolved organisms from one of the runs is not equivalent to taking ten organisms from ten different runs, as in the former case the top ten organisms from the same run will have mostly similar parameter values.

An alternative method for generating multiple evolved organisms would be to use different training sets for each run. For example, we might use a specific grandmaster for each run in the hope of obtaining organisms that mimic the styles of various grandmasters. However, our tests indicate that this approach does not improve over the method used. Apparently, using 1-ply searches only enables mimicking general grandmaster style, rather than the subtleties of a specific player.

In the coevolution phase, the ten best organisms selected serve as the initial population, which is then coevolved over 50 generations. In each generation, each organism plays four games against each other organism (to obtain a more reliable result). Note that for each game, a different sequence of opening moves is selected from an opening book file. This ensures that each game is unique (i.e., each game contains a unique set of opening moves). At the end of each generation, rank-based selection is applied for selecting the organisms for breeding. Elitism is used here as well, which ensures that the best organism survives for the next generation. This is especially critical in light of the small population size. Other GA parameters remain unchanged, that is, uniform crossover with crossover rate of 0.75 and mutation rate of 0.005.

Note that at this phase we still do not have a tuned search mechanism. So, for the coevolution phase described above, we use a basic alpha-beta search mechanism, without any of the advanced selective search mechanisms (for which we have yet to evolve their parameter values). As a result, the playing strength of the program at this stage is substantially limited. However, this does not pose a problem, since this limitation equally applies to all ten programs which play against each other.

In the following section we present our method for evolving the parameter values of the selective search mechanism.

\section{\label{sec:search-evolution}Evolution of Selective Search}

After evolving the parameter values of the evaluation function and incorporating the values of the best evolved organism into the program, we now focus on evolving the parameter values of the search mechanism.

Several popular selective search methods are employed by top tournament-playing programs. These methods allow the program to search more selectively, i.e., prune ``uninteresting'' moves earlier, thus spending additional time on more promising moves. That is, instead of searching all the moves to a certain fixed depth, some moves are searched more deeply than others.

The most popular selective search methods are null-move pruning \cite{beal89,david08b,donninger93}, futility pruning \cite{heinz98a}, multi-cut pruning \cite{bjornsson98,bjornsson01}, and selective extensions \cite{anantharaman91,beal95}. A description of these methods can be found in \cite{david10}. Each of these selective search methods requires several critical parameters to be tuned. Normally these parameters are manually tuned, usually through years of experiments and manual optimizations.

In order to apply GA for tuning the parameters of these selective search methods automatically, we represent these parameters as a binary chromosome where each parameter's number of allocated bits is based on the reasonable ranges for the values of the parameter. Table~\ref{tab:chromosome} presents the chromosome and the range of values for each parameter. Note that for search extensions the notion fractional ply is applied, where 1 ply $=$ 4 units (e.g., an extension value of 2 is equivalent to half a ply, etc.).

\begin{table}[htbp]
\begin{center}
\begin{tabular}{|l||c|c|}
\hline
Parameter & Value range & Bits\\
\hline
\hline
Null-move use & 0--1 & 1\\
\hline
Null-move reduction & 0--7 & 3\\
\hline
Null-move use adaptivity & 0--1 & 1\\
\hline
Null-move adaptivity depth & 0--7 & 3\\
\hline
Futility depth & 0--3 & 2\\
\hline
Futility threshold depth-1 & 0--1023 & 10\\
\hline
Futility threshold depth-2 & 0--1023 & 10\\
\hline
Futility threshold depth-3 & 0--1023 & 10\\
\hline
Multi-cut use & 0--1 & 1\\
\hline
Multi-cut reduction & 0--7 & 3\\
\hline
Multi-cut depth & 0--7 & 3\\
\hline
Multi-cut move num & 0--31 & 5\\
\hline
Multi-cut cut num & 0--7 & 3\\
\hline
Check extension & 0--4 & 3\\
\hline
One-reply extension & 0--4 & 3\\
\hline
Recapture extension & 0--4 & 3\\
\hline
Passed pawn extension & 0--4 & 3\\
\hline
Mate threat extension & 0--4 & 3\\
\hline
\hline
Total chromosome length & & 70\\
\hline
\end{tabular}
\end{center}
\caption{Chromosome representation of 18 search parameters (length: 70 bits)}
\label{tab:chromosome}
\end{table}

We employ the optimization goal in Bj\"ornsson and Marsland \cite{bjornsson02} as our fitness function. A set of 879 tactical test positions from Encyclopedia of Chess Middlegames (ECM) was used for training. Each position in the ECM test suite has a predetermined ``best move''. Each chromosome processes all of the 879 positions, and for each position it attempts to find this predetermined best move as fast as possible.

Instead of counting the number of correctly ``solved'' positions (number of positions for which the organism found the best move), we used the number of nodes the organism had to process in order to find the best move. For each position, we recorded the number of nodes the organism searched before finding the correct move. The total node count for each organism is the total node count for all the 879 positions. We imposed an upper search limit of 500,000 nodes per position. That is, if the correct move is not found after searching 500,000 nodes, the search is stopped and this upper limit is returned as the number of nodes for that position. Naturally, more positions will be solved if a larger search limit is chosen, but also more time will be spent, and subsequently the whole evolution will take much longer.

The lower the total node count for all the positions, the higher the fitness of the organism will be. Using this fitness measure instead of taking, in a straightforward manner, the number of solved positions has the benefit of deriving more fitness information per position. Rather than obtaining a 1-bit information (correct/false) for each position, a numeric value is obtained, which also measures how quickly the position is solved. Thus, the organism is not only ``encouraged'' to solve more positions, but also finding quicker solutions for the already solved test positions.

Similarly to \cite{david10}, we used a standard GA implementation with Gray coded chromosomes, fitness-proportional selection, uniform crossover, and elitism. All the organisms are initialized with random values. The following parameters are used for the GA:
\\
\\
population size = 10,\\
crossover rate = 0.75,\\
mutation rate = 0.05,\\
number of generations = 50.\\

The next section contains the experimental results using the GA-based method for the parameters of the evaluation function and the search mechanism.

\section{\label{sec:experiments}Experimental Results}

We now present the results of running the evolutionary process described in the previous two sections. We also provide the results of several experiments that measure the strength of the evolved program in comparison to \textsc{Crafty}, a former two-time World Computer Chess Champion that is commonly used as a baseline for testing chess programs.

\subsection{Results of Evolution}

\begin{figure}[htbp]
\centering
\includegraphics[height=2.2in, width=3.1in]{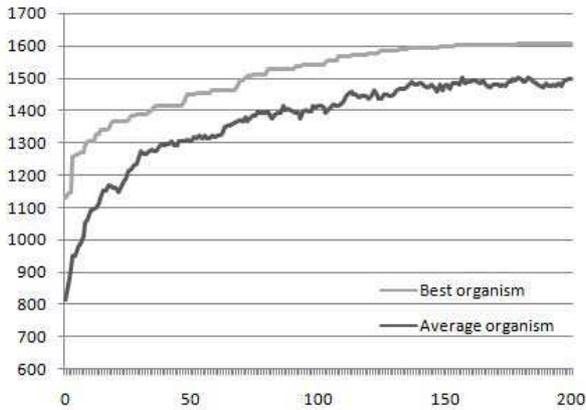}
\caption{Number of positions solved (out of 5,000) for the best organism and the population average in each generation (total time for 200 generations $\approx$ 2 hours). The figure shows the results for the first run, out of a total of ten runs.}
\label{fig:graph}
\end{figure}

Figure~\ref{fig:graph} shows the number of positions solved (i.e., the number of correct moves found) for the best organism and the population average for 200 generations (the figure shows the results for the first run, out of a total of ten runs). Specifically, the results indicate that the average number of solved positions is about 800 (out of 5,000) in the first generation. Note that even without any chess knowledge an organism would occasionally select the correct move by random guessing. Additionally, since the randomly initialized parameters contain only positive values, an organism can find some basic captures of the opponent's pieces without possessing any real chess knowledge. 

The average number of solved positions increases until stabilizing at around 1500, which corresponds to 30\% of the positions. The best organism at generation 200 solves 1620 positions, which corresponds to 32.4\% of the positions. Due to the use of elitism, the number of solved positions for the best organism is monotonously increasing, since the best organism is preserved. The entire 200-generation evolution took approximately 2 hours.

At first glance, a solution rate of 32\% might not seem too high. However, in light of the fact that the organisms base their move on a 1-ply search only (as opposed to the thorough analysis and consideration of the position by a grandmaster prior to their move), this figure is quite satisfactory. In other words, by conducting merely a 1-ply search, the evolved organism selects successfully a grandmaster's carefully analyzed move in one out of three cases. 

With the completion of the learning phase, we used the additional 5,000 positions set aside for testing. We let our best evolved organism perform a 1-ply search on each of these positions. The number of correctly solved positions was 1521 (30.4\%). This indicates that the first 5,000 positions used for training cover most types of positions that can arise, as the success rate for the testing set is close to the success rate for the training set.

\subsection{Results of Coevolution}

Repeating the evolutionary process, we obtained each time a ``best evolved organism'' with a different set of evolved parameter values. That is, each run produced a different grandmaster-level program. Even though the  performance of these independently evolved best organisms is fairly similar, our goal was to improve upon these organisms and create an enhanced ``best of best'' organism.

\begin{figure}[htbp]
\begin{center}
\line(1,0){240}
\begin{verbatim}
  PAWN_VALUE                         100
  KNIGHT_VALUE                       521
  BISHOP_VALUE                       572
  ROOK_VALUE                         824
  QUEEN_VALUE                       1710
  PAWN_ADVANCE_A                       3
  PAWN_ADVANCE_B                       6
  PASSED_PAWN_MULT                    10
  DOUBLED_PAWN_PENALTY                14
  ISOLATED_PAWN_PENALTY                8
  BACKWARD_PAWN_PENALTY                3
  WEAK_SQUARE_PENALTY                  5
  PASSED_PAWN_ENEMY_KING_DIST          7
  KNIGHT_SQ_MULT                       6
  KNIGHT_OUTPOST_MULT                  9
  BISHOP_MOBILITY                      4
  BISHOP_PAIR                         28
  ROOK_ATTACK_KING_FILE               51
  ROOK_ATTACK_KING_ADJ_FILE            8
  ROOK_ATTACK_KING_ADJ_FILE_ABGH      26
  ROOK_7TH_RANK                       30
  ROOK_CONNECTED                       6
  ROOK_MOBILITY                        4
  ROOK_BEHIND_PASSED_PAWN             40
  ROOK_OPEN_FILE                      27
  ROOK_SEMI_OPEN_FILE                 11
  ROOK_ATCK_WEAK_PAWN_OPEN_COLUMN     15
  ROOK_COLUMN_MULT                     6
  QUEEN_MOBILITY                       2
  KING_NO_FRIENDLY_PAWN               35
  KING_NO_FRIENDLY_PAWN_ADJ           10
  KING_FRIENDLY_PAWN_ADVANCED1         6
  KING_NO_ENEMY_PAWN                  17
  KING_NO_ENEMY_PAWN_ADJ               9
  KING_PRESSURE_MULT                   4
\end{verbatim}
\line(1,0){240}
\caption{Average evolved parameters of the evaluation function of the best individual after ten runs (the values are rounded).}
\label{fig:best-evolved}
\end{center}
\end{figure}

We applied single-population coevolution to enhance the performance of the program. After running the evolution ten times (which ran for about 20 hours), ten different best organisms were obtained. Using these ten organisms as the starting population, we applied GA for 50 generations, where each organism played each other organism four times in every round. Each game was limited to ten seconds (5 seconds per side). In practice, this coevolution phase ran for approximately 20 hours.

Figure~\ref{fig:best-evolved} provides the average evolved values of the best individual obtained after running our method ten times (the values are rounded). Conventionally, knight and bishop are valued at 3 pawns, and rook and queen are valued at 5 and 9 pawns, respectively \cite{capablanca06}. These values are used in most chess programs. However, it is well known that grandmasters frequently sacrifice pawns for positional advantages, and that in practice, a pawn is assigned a lower value. Interestingly, the best organism assigns a pawn about half the value it is usually assigned, relative to the other pieces, which is highly unconventional for chess programs. This implies that the evolved organism, which learns its parameter values from human grandmasters, ends up adopting also their less materialistic style of play. This is also reflected in the playing style of the ultimate evolved program, as it frequently sacrifices pawns for gaining positional advantages.

\subsection{Evolving Search Parameters}

\begin{figure}[htbp]
\centering
\includegraphics[height=2.04in, width=3.2in]{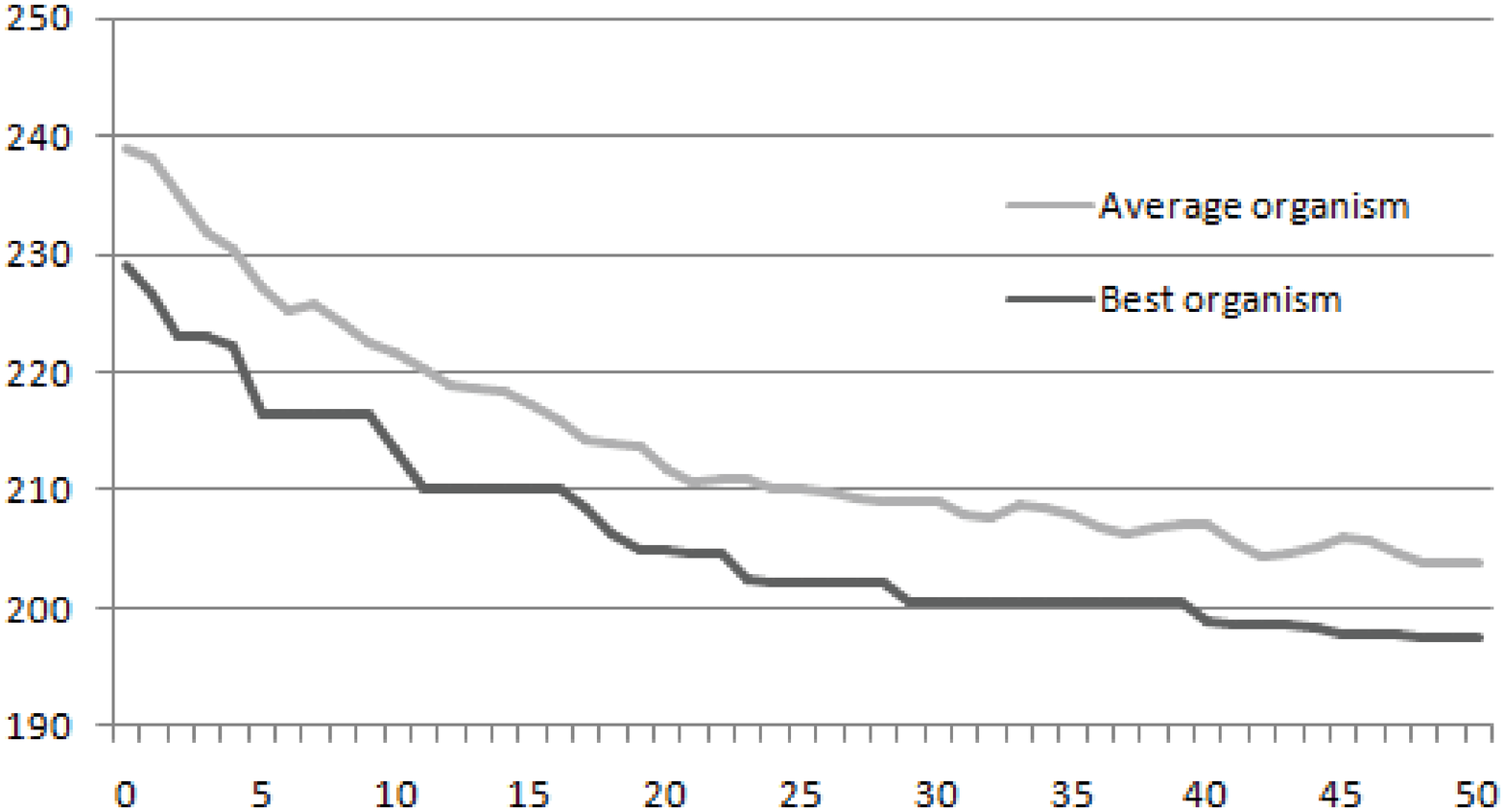}
\caption{Total node count (in millions) for 879 ECM positions for the best organism and the population average in each generation.}
\label{fig:search-graph}
\end{figure}

After incorporating the best evolved parameters for the evaluation function into the program, we then evolved the search parameters. Figure~\ref{fig:search-graph} shows the total node count for 879 positions for the best organism and the population average in each generation. Table~\ref{tab:chromosome-val} provides the evolved values of the best individual.

\begin{table}[htbp]
\begin{center}
\begin{tabular}{|l|c|}
\hline
Parameter & Learned value\\
\hline
\hline
Null-move use & 1\\
\hline
Null-move reduction & 4\\
\hline
Null-move use adaptivity & 1\\
\hline
Null-move adaptivity depth & 6\\
\hline
Futility depth & 3\\
\hline
Futility threshold depth-1 & 112\\
\hline
Futility threshold depth-2 & 227\\
\hline
Futility threshold depth-3 & 506\\
\hline
Multi-cut use & 1\\
\hline
Multi-cut reduction & 4\\
\hline
Multi-cut depth & 6\\
\hline
Multi-cut move num & 15\\
\hline
Multi-cut cut num & 3\\
\hline
Check extension & 4\\
\hline
One-reply extension & 4\\
\hline
Recapture extension & 2\\
\hline
Passed pawn extension & 3\\
\hline
Mate threat extension & 2\\
\hline
\end{tabular}
\end{center}
\caption{Learned values for the search parameters}
\label{tab:chromosome-val}
\end{table}

We incorporate the values obtained here together with the values obtained from evolution of the parameter values of the evaluation function into our program (which is an experimental descendant of the program \textsc{Falcon} \cite{david05,david09a}). We call this program \textsc{Evol*}. The program contains all our obtained parameter values for the evaluation function and the search mechanism.

To measure the performance of \textsc{Evol*}, we conducted a series of matches against the chess program \textsc{Crafty} \cite{hyatt90}. \textsc{Crafty} has successfully participated in numerous World Computer Chess Championships (WCCC), and is a direct descendent of \textsc{Cray Blitz}, the WCCC winner of 1983 and 1986. It is frequently used in the literature as a standard reference. In order to obtain a measure for the performance gain due to the coevolution phase, we also compared the performance of \textsc{Evol*} to a version of \textsc{Evol*} which uses the parameters in its evaluation function that are evolved after the evolutionary phase (i.e., parameters evolved before the coevolution phase). We call this program \textsc{Evol$_0$}. For the search mechanism we used the same parameter setting in \textsc{Evol*} and \textsc{Evol$_0$}.

%------------------------------
\begin{table}[htbp]
\begin{center}
\begin{tabular}{|c|c|c|}
\hline
\textsc{Evol*} & \textsc{Evol$_0$} & \textsc{Crafty}\\
\hline
\hline
654 & 623 & 593\\
\hline
\end{tabular}
\end{center}
\caption{Number of ECM positions solved by each program (time: 5 seconds per position)}
\label{tab:search-ecm}
\end{table}
%------------------------------

For a proper comparison, we let \textsc{Evol*}, \textsc{Evol$_0$}, and \textsc{Crafty} process the ECM test suite with 5 seconds per position. Table~\ref{tab:search-ecm} provides the results. As can be seen, both \textsc{Evol*} and \textsc{Evol$_0$} solve significantly more positions than \textsc{Crafty}.

The superior performance of \textsc{Evol*} on the ECM test set is not surprising, as it was evolved on this training set. Therefore, in order to obtain an unbiased performance comparison, we conducted a series of 300 matches against \textsc{Crafty}, using a time control of 5 minutes per game for each side. Table~\ref{tab:search-matches} provides the results.

%------------------------------
\begin{table}[htbp]

\begin{center}
\begin{tabular}{|l||c|c|c|}
\hline
Match & Result & W\% & RD\\
\hline
\hline
\textsc{Evol*} - \textsc{Evol$_0$} & 180.0 - 120.0 & 60.0\% & $+70.4$\\
\hline
\textsc{Evol*} - \textsc{Crafty} & 181.5 - 118.5 & 60.5\% & $+74.1$\\
\hline
\end{tabular}
\end{center}
\caption{\textsc{Evol*} vs.~\textsc{Evol$_0$} and \textsc{Crafty} (W\% is the winning percentage, and RD is the Elo rating difference). Time control: 5 minutes per game.}
\label{tab:search-matches}

\end{table}
%------------------------------

The results further show that \textsc{Evol*} outperforms (\textsc{Evol$_0$} and) \textsc{Crafty}, not only in terms of solving more tactical test positions, but more importantly in test matches. These results establish that although the search parameters were evolved from randomly initialized chromosomes, the resulting organism outperforms a grandmaster-level chess program in 5 minute games.

We extended our experiments to compare the performance of \textsc{Evol*} against several of the world's top commercial chess programs. These programs included \textsc{Junior}, \textsc{Fritz}, and \textsc{Hiarcs}. \textsc{Junior} won the World Microcomputer Chess Championship in 1997 and 2001 and the World Computer Chess Championship in 2002, 2004, 2006, and 2011. In 2003 \textsc{Junior} played a 6-game match against the legendary former world champion, Garry Kasparov, that resulted in a 3--3 tie. In 2007 \textsc{Junior} won the ``ultimate computer chess challenge'' organized by the World Chess Federation (FIDE), defeating \textsc{Fritz} 4--2. \textsc{Fritz} won the World Computer Chess Championship in 1995. In 2002 it drew the ``Brains in Bahrain'' match against former world champion, Vladimir Kramnik, 4?-4, and in 2003 it drew again a four-game match against Garry Kasparov. In 2006 \textsc{Fritz} defeated the incumbent world champion, Vladimir Kramnik, 4--2. \textsc{Hiarcs} won the 1993 World Microcomputer Chess Championship. In 2003 it drew a four-game match against Grandmaster Evgeny Bareev, then the 8th ranked player in the world. (All four games ended in a draw.) In 2007 \textsc{Hiarcs} won the 17th International Paderborn Computer Chess Championship.

As mentioned earlier, \textsc{Evol*} is an experimental descendant of the program \textsc{Falcon} \cite{david05,david09a}, which successfully participated in three World Computer Chess Championships. During the 2008 World Computer Chess Championship \cite{david08}, \textsc{Falcon} used an earlier version \cite{david11} of the evolutionary approach described in this paper. Competing with an average laptop against nine strong chess programs (eight of which ran on fast multicore machines ranging from 4 to 40 cores), the GA-based version of \textsc{Falcon} reached second place in the World Computer Speed Chess Championship and sixth place in the World Computer Chess Championship. These highly surprising results, especially in light of the huge hardware handicap of \textsc{Falcon} relatively to its competitors, demonstrate the capabilities of the GA-based approach.

Table~\ref{tab:commercial} provides the results of \textsc{Evol*} against the top commercial programs \textsc{Junior}, \textsc{Fritz}, and \textsc{Hiarcs}. Again, all matches consisted of 300 games at a time control of 5 minutes per game for each side.

%------------------------------
\begin{table}[htbp]

\begin{center}
\begin{tabular}{|l||c|c|c|}
\hline
Match & Result & W\% & RD\\
\hline
\hline
\textsc{Evol*} - \textsc{Junior} & 133.5 - 166.5 & 44.5\% & $-38.4$\\
\hline
\textsc{Evol*} - \textsc{Fritz} & 162.0 - 138.0 & 54.0\% & +27.9\\
\hline
\textsc{Evol*} - \textsc{Hiarcs} & 180.5 - 119.5 & 60.2\% & +71.6\\
\hline
\end{tabular}
\end{center}
\caption{\textsc{Evol*} vs.~\textsc{Junior}, \textsc{Fritz}, and \textsc{Hiarcs} (W\% is the winning percentage, and RD is the Elo rating difference). Time control: 5 minutes per game.}
\label{tab:commercial}

\end{table}
%------------------------------

To further examine the performance of \textsc{Evol*} under long time control conditions, we conducted a series of matches between \textsc{Evol*} and the abovementioned top tournament-playing programs at a time control of 1 hour per game for each side. Each match consisted of 100 games. Table~\ref{tab:commercial-long} provides the results.

%------------------------------
\begin{table}[htbp]

\begin{center}
\begin{tabular}{|l||c|c|c|}
\hline
Match & Result & W\% & RD\\
\hline\hline
\textsc{Evol*} - \textsc{Crafty} & 61.5 - 38.5 & 61.5\% & $+81.4$\\
\hline
\textsc{Evol*} - \textsc{Junior} & 41.0 - 59.0 & 41.0\% & $-63.2$\\
\hline
\textsc{Evol*} - \textsc{Fritz} & 51.5 - 48.5 & 51.5\% & +10.4\\
\hline
\textsc{Evol*} - \textsc{Hiarcs} & 58.5 - 41.5 & 58.5\% & +59.6\\
\hline
\end{tabular}
\end{center}
\caption{\textsc{Evol*} vs.~\textsc{Crafty}, \textsc{Junior}, \textsc{Fritz}, and \textsc{Hiarcs} (W\% is the winning percentage, and RD is the Elo rating difference). Time control: 1 hour per game.}
\label{tab:commercial-long}

\end{table}
%------------------------------

The results show that the performance of a genetically evolved program is on a par with that of the top commercial chess programs. The superior performance by \textsc{Junior} is well noted and should serve as a source of inspiration. In addition, Table~\ref{tab:ecm2} compares the tactical performance of our evolved organism against these three commercial programs. The results show the number of ECM positions solved by each program. A similar trend emerges, i.e., the evolved organism is on a par with these top commercial programs.

%------------------------------
\begin{table}[htbp]
\begin{center}
\begin{tabular}{|c|c|c|c|}
\hline
\textsc{Evol*} & \textsc{Junior} & \textsc{Fritz} & \textsc{Hiarcs}\\
\hline
\hline
654 & 681 & 640 & 642\\
\hline
\end{tabular}
\end{center}
\caption{Number of ECM positions solved by each program (time: 5 seconds per position)}
\label{tab:ecm2}
\end{table}
%------------------------------

The results of the abovementioned tests establish that even though the parameters of our program are evolved from chromosomes initialized randomly, the resulting organism is on a par with top commercial chess programs.

%\newpage
\section{\label{sec:conclusions}Concluding Remarks}

In this paper we presented a novel approach for evolving the key components of a chess program from randomly initialized values using genetic algorithms.

In contrast to our previous successful attempts which focused on mimicking the evaluation function of a chess program acting as a mentor, the approach presented in this paper focuses on evolving the parameters of the evaluation function and the search mechanism. It is done by observing solely (human) grandmaster-level games, where the only available information to guide the evolution is the moves played in these games.

Learning from the actual moves of grandmaster-level games in the first phase of the evolution, we obtained several evaluation functions. Specifically, running the procedure ten times, we obtained ten such evolved evaluation functions, which served as the initial population for the second coevolution phase. Using coevolution in our case proved successful as the initial population was not random, but relatively well tuned due to the first phase of the evolution.

We further used genetic algorithms to evolve the parameters of the search mechanism. Starting from randomly initialized values, our method evolves these parameters. Combining the two parameter sets (i.e., those of the evaluation function and the search mechanism) resulted in a performance that is on a par with that of top tournament-playing programs.

Overall, this paper provides, to the best of our knowledge, the first methodology of automatic learning the parameters of the evaluation function and the search mechanism from randomly initialized values for computer chess. We note that although the experimental results of the module presented here differ from those of our previous work (which assumed access to a highly tuned evaluation function \cite{david10} or a highly tuned search mechanism \cite{david09,david11}), the overall performance observed was comparable. Thus, the approach in this paper is superior to these previous efforts, in the sense that it achieves grandmaster-level performance without relying on any assumptions.

The results presented in this paper point to the vast potential in applying evolutionary methods for learning from human experts. We believe that this approach could be applied to a wide array of problems for essentially ``reverse engineering'' the knowledge of a human expert. While we successfully used genetic algorithms to evolve the parameter values of a chess program, it would be of interest to apply other natural and evolutionary optimization methods in future research (e.g., covariance matrix adaptation evolution strategy (CMA-ES) \cite{hansen01}, differential evolution (DE) \cite{storn97}, and particle swarm optimization (PSO) \cite{kennedy95}).

%\newpage

\appendices

\section{Experimental Setup}

\hspace*{-4pt}Our experimental setup consisted of the following resources:

\begin{itemize}
\item \textsc{Crafty}, \textsc{Junior}, \textsc{Fritz}, and \textsc{Hiarcs} chess programs running as native ChessBase engines.

\item Encyclopedia of Chess Middlegames (ECM) test suite, consisting of 879 positions.

\item \textsc{Fritz 9} interface for automatic running of matches, using \textsc{Shredder} opening book.

\item Long time control matches (Table~\ref{tab:commercial-long}) were conducted on an Intel Core 2 Quad Q8300 with 8.0 GB RAM. All the other experiments were conducted on an Intel Core 2 Duo T8100 with 3.0 GB RAM. 
\end{itemize}

\section{Elo Rating System}

The Elo rating system, developed by Arpad E.~Elo \cite{elo78}, is the official system for calculating the relative skill levels of players in chess. A grandmaster is generally associated with rating values of above 2500 Elo, and a novice player with rating values of below 1400 Elo.

Given the rating difference ($RD$) of two players, the following formula calculates the expected winning rate ($W$, between 0 and 1) of the player:

\begin{displaymath}
W = \frac{1}{10^{-RD/400} + 1}.
\end{displaymath}

Given the winning rate of a player, as is the case in our experiments, the expected rating difference can be derived from the above formula:

\begin{displaymath}
RD = -400 \log_{10}(\frac{1}{W} - 1).
\end{displaymath}

%-------------------------------

\end{document}